\documentclass{article}

 \usepackage[nonatbib,final]{nips_2018}

\usepackage[utf8]{inputenc} 
\usepackage[T1]{fontenc}    
\usepackage{hyperref}       
\usepackage{url}            
\usepackage{booktabs}       
\usepackage{multirow}
\usepackage{amsfonts}       
\usepackage{nicefrac}       
\usepackage{microtype}      
\usepackage{graphicx}
\usepackage{amsmath,amssymb} 
\usepackage{color}
\usepackage{cite}

\title{Pairwise Relational Networks using Local Appearance Features for Face Recognition}

\author{
  Bong-Nam Kang \\
  Department of Creative IT Engineering\\
  POSTECH, Korea\\
  \texttt{bnkang@postech.ac.kr} \\
  \And
  Yonghyun Kim, Daijin Kim \\
  Department of Computer Science and Engineering \\
  POSTECH, Korea\\
 \texttt{\{gkyh0805,dkim\}@postech.ac.kr} \\
}

\begin{document}

\maketitle

\begin{abstract}
 We propose a new face recognition method, called a pairwise relational network (PRN), which takes local appearance features around landmark points on the feature map, and captures unique pairwise relations with the same identity and discriminative pairwise relations between different identities. The PRN aims to determine facial part-relational structure from local appearance feature pairs. Because meaningful pairwise relations should be identity dependent, we add a face identity state feature, which obtains from the long short-term memory (LSTM) units network with the sequential local appearance features. To further improve accuracy, we combined the global appearance features with the pairwise relational feature. Experimental results on the LFW show that the PRN achieved $99.76\%$ accuracy. On the YTF, PRN achieved the \textit{state-of-the-art} accuracy ($96.3\%$). The PRN also achieved comparable results to the \textit{state-of-the-art} for both face verification and face identification tasks on the IJB-A and IJB-B. This work is already published on ECCV 2018.
\end{abstract}

\section{Introduction}
Face recognition in unconstrained environments is a very challenging problem in computer vision. Faces of the same identity can look very different when presented in different illuminations, facial poses, facial expressions, and occlusions. Such variations within the same identity could overwhelm the variations due to identity differences and make face recognition challenging. To solve these problems, many deep learning-based approaches have been proposed and achieved high accuracies of face recognition such as DeepFace \cite{DeepFace}, DeepID series \cite{DeepID,DeepID2,DeepID2+,DeepID3}, FaceNet \cite{FaceNet},  PIMNet \cite{PIMNet_CVPRW2017}, SphereFace \cite{SphereFace}, and ArcFace \cite{ArcFace}.
In face recognition tasks in unconstrained environments, the deeply learned and embedded features need to be not only separable but also discriminative. However, these features are learned implicitly for separable and distinct representations to classify between different identities without what part of the features is used, what part of the feature is meaningful, and what part of the features is separable and discriminative. Therefore, it is difficult to know what kinds of features are used to discriminate the identities of face images clearly.
To overcome this limitation, we propose a novel face recognition method, called a pairwise relational network (PRN) to capture unique relations within same identity and discriminative relations between different identities. To capture relations, the PRN takes local appearance features as input by ROI projection around landmark points on the feature map. With these local appearance features, the PRN is trained to capture unique pairwise relations between pairs of local appearance features to determine facial part-relational structures and properties in face images. Because the existence and meaning of pairwise relations should be identity dependent, the PRN could condition its processing on the facial identity state feature. The facial identity state feature is learned from the long short-term memory (LSTM) units network with the sequential local appearance features on the feature maps. To more improve accuracy of face recognition, we combined the global appearance features with the relation features. 
We present extensive experiments on the public available datasets such as Labeled Faces in the Wild (LFW) \cite{LFW}, YouTube Faces (YTF) \cite{YTF_CVPR2011}, IARPA Janus Benchmark A (IJB-A) \cite{IJB-A_CVPR2015}, and IARPA Janus Benchmark B (IJB-B) \cite{IJB-B_CVPRW2017} and show that the proposed PRN is very useful to enhance the accuracy of both face verification and face identification.

\section{Pairwise relational network}\label{sec:PRN}
The pairwise relational network (RRN) takes a set of local appearance features on the feature map as its input and outputs a single vector as its relational representation feature for the face recognition task. The PRN captures unique and discriminative pairwise relations between different identities. In other words, the PRN captures the core unique and common properties of faces within the same identity, whereas captures the separable and discriminative properties of faces between different identities. Therefore, the PRN aims to determine pairwise-relational structures from pairs of local appearance features in face images. The relational feature $\boldsymbol{r}_{i,j}$ represents a latent relation of a pair of two local appearance features, and can be written as follows:
\begin{equation}
	\boldsymbol{r}_{i,j} = \mathcal{G}_{\theta}\left(\boldsymbol{p}_{i,j}\right),
	\label{eq:eq_relation}
\end{equation}
where $\mathcal{G}_{\theta}$ is a multi-layer perceptron (MLP) and its parameters $\theta$ are learnable weights. $\boldsymbol{p}_{i,j} = \{\boldsymbol{f}^{l}_{i}, \boldsymbol{f}^{l}_{j}\}$ is a pair of two local appearance features, $\boldsymbol{f}^{l}_{i}$ and $\boldsymbol{f}^{l}_{j}$, which are $i$-th and $j$-th local appearance features corresponding to each facial landmark point, respectively. Each $\boldsymbol{f}^{l}_{i}$ is extracted by the RoI projection which projects a $m\times m$ region around $i$-th landmark point in the input facial image space to a $m^{'}\times m^{'}$ region on the feature maps space. The same MLP operates on all possible parings of local appearance features.
\begin{figure}[t]
	\centering
	\includegraphics[scale=0.33]{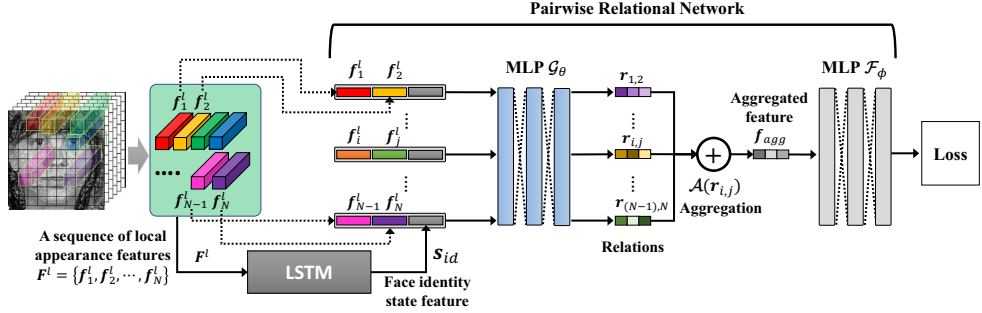}
	\caption{Pairwise Relational Network. The PRN captures unique and discriminative pairwise relations dependent on facial identity.}
	\label{fig:fig_PRN_sid}
\end{figure}

However, the permutation order of local appearance features is a critical for the PRN, since without this invariance, the PRN would have to learn to operate on all possible permuted pairs of local appearance features without explicit knowledge of the permutation invariance structure in the data. To incorporate this permutation invariance, we constrain the PRN with an aggregation function (\figurename~\ref{fig:fig_PRN_sid}):
\begin{equation}
	\boldsymbol{f}_{agg} = \mathcal{A}(\boldsymbol{r}_{i,j}) = \sum_{\forall \boldsymbol{r}_{i,j}}{\left(\boldsymbol{r}_{i,j}\right)},
	\label{eq:eq_aggregation}
\end{equation}
where $\boldsymbol{f}_{agg}$ is the aggregated relational feature, and $\mathcal{A}$ is the aggregation function which is summation of all pairwise relations among all possible pairing of the local appearance features.
Finally, a prediction $\widetilde{\boldsymbol{r}}$ of the PRN can be performed with $\widetilde{\boldsymbol{r}}= \mathcal{F}_{\phi}\left(\boldsymbol{f}^{agg}\right),\label{eq:eq_prediction}$ where $\mathcal{F}_{\phi}$ is a function with parameters $\phi$, and are implemented by the MLP. Therefore, the final form of the PRN is a composite function as follows:
\begin{equation}
	PRN(\boldsymbol{P}) = \mathcal{F}_{\phi}\left(\mathcal{A}\left(\mathcal{G}_{\theta}\left(\boldsymbol{p}_{i,j}\right)\right)\right),
	\label{eq:eq_prn_rev}
\end{equation}
where $\boldsymbol{P} = \{\boldsymbol{p}_{1,2}, \ldots, \boldsymbol{p}_{i,j}, \ldots, \boldsymbol{p}_{(N-1), N}\}$ is a set of all possible pairs of local appearance features where $N$ denotes the number of local appearance features on the feature maps.

To capture unique and discriminative pairwise relations among different identities, a pairwise relation should be identity dependent. So, we modify the PRN such that $\mathcal{G}_{\theta}$ could condition its processing on the identity information. To condition the identity information, we embed a face identity state feature $\boldsymbol{s}_{id}$ as the identity information in the $PRN$:
\begin{equation}
	PRN^{+}(\boldsymbol{P}, \boldsymbol{s}_{id}) = \mathcal{F}_{\phi}\left(\mathcal{A}\left(\mathcal{G}_{\theta}\left(\boldsymbol{p}_{i,j}, \boldsymbol{s}_{id}\right)\right)\right).
	\label{eq:equation_prn_idf}
\end{equation}
To get this $\boldsymbol{s}_{id}$, we use the final state of a recurrent neural network composed of LSTM layers and two fully connected layers that process a sequence of local appearance features: $\boldsymbol{s}_{id}=\mathcal{E}_{\psi}(\boldsymbol{F}^{l}),$ where $\mathcal{E}_{\psi}$ is a neural network module which composed of LSTM layers and two fully connected layers with learnable parameters $\psi$. We train $\mathcal{E}_{\psi}$ with \textit{softmax} loss function. The detailed configuration of $\mathcal{E}_{\psi}$ used in our proposed method is in Appendix \ref{sec:sid_config}.

\paragraph{Loss function}
To learn the PRN, we use jointly the triplet ratio loss $L_{t}$, pairwise loss $L_{p}$, and \textit{softmax} loss $L_{s}$ to minimize distances between faces that have the same identity and to maximize distances between faces that are of different identity. 
$L_{t}$ is defined to maximize the ratio of distances between the positive pairs and the negative pairs in the triplets of faces $T$. To maximize $L_{t}$, the Euclidean distances of positive pairs should be minimized and those of negative pairs should be maximized. Let $F(I)\in\mathbb{R}^{d}$, where $I$ is the input facial image, denote the output of a network, the $L_{t}$ is defined as follows:
\begin{equation}
	L_{t} = \sum_{\forall T}\max\left(0, 1 - \frac{\|F(I_{a}) - F(I_{n})\|_{2}}{\left\|F(I_{a}) - F(I_{p})\right\|_{2} + m}\right),
	\label{eq:eq_triplet_ratio_loss}
\end{equation}
where $F(I_{a})$ is the output for an anchor face $I_{a}$, $F(I_{p})$ is the output for a positive face image $I_{p}$, and $F(I_{n})$ is the output for a negative face $I_{n}$ in ${T}$, respectively. $m$ is a margin that defines a minimum ratio in Euclidean space. From recent work by Kang \textit{et al}. \cite{PIMNet_CVPRW2017}, they reported that although the ratio of the distances is bounded in a certain range of values, the range of the absolute distances is not. To solve this problem, they constrained $L_{t}$ by adding the pairwise loss function $L_{p}$.
$L_{p}$ is defined to minimize the sum of the squared Euclidean distances between $F(I_{a})$ for the anchor face and $F(I_{p})$ for the positive face. These pairs of $I_{a}$ and $I_{p}$ are in the triplets of faces $T$.
\begin{equation}
L_{p} = \sum_{(I_{a}, I_{p}) \in T}\|F(I_{a}) - F(I_{p}) \|_{2}^{2}.
\end{equation}
The joint training with $L_{t}$ and $L_{p}$ minimizes the absolute Euclidean distance between face images of a given pair in the triplets of facs $T$. We also use these loss functions with \textit{softmax} loss $L_{s}$ jointly.

\section{Experiments}\label{sec:expt}
We evaluated the proposed face recognition method on the public available benchmark datasets such as the LFW, YTF, IJB-A, and IJB-B. For fair comparison in terms of the effects of each network module, we train three kinds of models (\textbf{model A} (base model, just use the global appearance feature $\boldsymbol{f}^{g}$), \textbf{model B} ($\boldsymbol{f}^{g}$ + $PRN$ in Eq, \eqref{eq:eq_prn_rev}), and \textbf{model C} ($\boldsymbol{f}^{g}$ + $PRN^{+}$ in Eq. \eqref{eq:equation_prn_idf}) under the supervision of cross-entropy loss with \textit{softmax} \cite{PIMNet_CVPRW2017}. More detailed configuration of them is presented in Appendix \ref{sec:detail_setting_models}.

\paragraph{Effects of the PRN and the face identity state feature}\label{sec:exp_prn}
To investigate the effectiveness of the PRN model with $\boldsymbol{s}_{id}$, we performed experiments in terms of the accuracy of classification on the validation set during training. For these experiments, we trained $PRN$ (Eq. \eqref{eq:eq_prn_rev}) and $PRN^{+}$ (Eq. \eqref{eq:equation_prn_idf}) with the face identity state feature $\boldsymbol{s}_{id}$. We achieved $94.2\%$ and $96.7\%$ accuracies of classification for $PRN$ and $PRN^{+}$, respectively. From this evaluation, when using $PRN^{+}$, we observed that the face identity state feature $\boldsymbol{s}_{id}$ represents the identity property, and the pairwise relations should be dependent on an identity property of a face image. Therefore, this evaluation validated the effectiveness of using the PRN network model and the importance of the face identity state feature.

\paragraph{Experiments on the Labeled Faces in the Wild (LFW)}\label{sec:exp_lfw}
From the experimental results on the LFW (See \tablename~\ref{tab:lfw_results} in Appendix \ref{sec:lfw_appendix}), we have the following observation.
First, \textbf{model C} (jointly combined $\boldsymbol{f}^{g}$ with $PRN^{+}$) beats the baseline model \textbf{model A} (the base CNN model, just uses $\boldsymbol{f}^{g}$) by significantly margin, improving the accuracy from $99.6\%$ to $99.76\%$. This shows that combination of $\boldsymbol{f}^{g}$ and $PRN^{+}$ can notably increase the discriminative power of deeply learned features, and the effectiveness of the pairwise relations between facial local appearance features.
Second, compared to \textbf{model B} (jointly combined $\boldsymbol{f}^{g}$ with $PRN$), \textbf{model C} achieved better accuracy of verification ($99.65\%$ \textit{vs.} $99.76\%$). This shows the importance of the face identity state feature $\boldsymbol{s}_{id}$ to capture unique and discriminative pairwise relations in the designed PRN model.
Last, compared to the \textit{state-of-the-art} methods on the LFW, the proposed \textbf{model C} is among the top-ranked approaches, outperforming most of the existing results (\tablename~\ref{tab:lfw_results} in Appendix \ref{sec:lfw_appendix}). This shows the importance and advantage of the proposed method.

\paragraph{Experiments on the YouTube Face Dataset (YTF)}\label{sec:exp_ytf}
From the experimental results on the YTF (See \tablename~\ref{tab:YTF_results} in Appendix \ref{sec:ytf_appendix}), we have the following observations.
First, \textbf{model C} beats the baseline model \textbf{model A} by a significantly margin, improving the accuracy from $95.1\%$ to $96.3\%$. This shows that combination of $\boldsymbol{f}^{g}$ and $PRN^{+}$ can notably increase the discriminative power of deeply learned features, and the effectiveness of the pairwise relations between local appearance features.
Second, compared to \textbf{model B}, \textbf{model C} achieved better accuracy of verification ($95.7\%$ \textit{vs.} $96.3\%$). This shows the importance of the face identity state feature $\boldsymbol{s}_{id}$ to capture unique and discriminative pairwise relations in the designed PRN model.
Last, compared to the \textit{state-of-the-art} methods on the YTF, the proposed method \textbf{model C} is the state-of-the-art ($96.3\%$ accuracy), outperforming the existing results (\tablename~\ref{tab:YTF_results} in Appendix \ref{sec:ytf_appendix}). This shows the importance and advantage of the proposed method.

\paragraph{Experiments on the IARPA Janus Benchmark A (IJB-A)}\label{sec:exp_ijb-a}
From the experimental results (See \tablename~\ref{tab:ijb-a_results} in Appendix \ref{sec:ijb-a_appendix}), we have the following observations.
First, compared to \textbf{model A}, \textbf{model C} achieves a consistently superior accuracy (TAR and TPIR) on both 1:1 face verification and 1:N face identification 
Second, compared to \textbf{model B}, \textbf{model C} achieved also a consistently better accuracy (TAR and TPIR) on both 1:1 face verification and 1:N face identification 
Last, more importantly, \textbf{model C} is trained from scratch and achieves comparable results to the \textit{state-of-the-art} (VGGFace2 \cite{VGG2Face}) which is first pre-trained on the MS-Celeb-1M dataset \cite{MS-Celeb-1M}, which contains roughly 10M face images, and then is fine-tuned on the VGGFace2 dataset. It shows that our proposed method can be further improved by training on the MS-Celeb-1M and fine-tuning our training dataset.

\paragraph{Experiments on the IARPA Janus Benchmark B (IJB-B)}\label{sec:exp_ijb-b}
From the experimental results (See \tablename~\ref{tab:ijb-b_results} in Appendix \ref{sec:ijb-b_appendix}), we have the following observations.
First, compared to \textbf{model A}, \textbf{model C} (jointly combined $\boldsymbol{f}^{g}$ with $PRN^{+}$ as the local appearance representation) achieved a consistently superior accuracy (TAR and TPIR) on both 1:1 face verification and 1:N face identification.
Second, compared to \textbf{model B} (jointly combined $\boldsymbol{f}^{g}$ with the $PRN$), \textbf{model C} achieved also a consistently better accuracy (TAR and TPIR) on both 1:1 face verification and 1:N face identification.
Last, more importantly, \textbf{model C} achieved consistent improvement of TAR and TPIR on both 1:1 face verification and 1:N face identification, and achieved the \textit{state-of-the-art} results on the IJB-B.

\section{Conclusion}\label{sec:conclusion}
We proposed a new face recognition method using the pairwise relational network (PRN) which takes local appearance feature around landmark points on the feature maps from the backbone network, and captures unique and discriminative pairwise relations between a pair of local appearance features. To capture unique and discriminative relations for face recognition, pairwise relations should be identity dependent. Therefore, the PRN conditioned its processing on the face identity state feature embedded by LSTM networks using a sequential local appearance features. To more improve accuracy of face recognition, we combined the global appearance feature with the PRN. Experiments verified the effectiveness and importance of our proposed PRN with the face identity state feature, which achieved $99.76\%$ accuracy on the LFW, the state-of-the-art accuracy ($96.3\%$) on the YTF, and comparable results to the state-of-the-art for both face verification and identification tasks on the IJB-A and IJB-B.

\section*{Acknoledgement}
This research was supported by the MSIT(Ministry of Science, ICT), Korea, under the SW Starlab support program (IITP-2017-0-00897), the “ICT Consilience Creative program” (IITP-2018-2011-1-00783), and “Development of Open Informal Dataset and Dynamic Object Recognition Technology Affecting Autonomous Driving” (IITP-2018-0-01290) supervised by the IITP.

\small
\bibliographystyle{IEEEtran}
\bibliography{egbib}

\clearpage
\appendix
\section{Implementation details}\label{sec:implement_details}
\subsection{Training data}
We used the web-collected face dataset (VGGFace2 \cite{VGG2Face}). All of the faces in the VGGFace2 dataset and their landmark points are detected by the recently proposed face detector \cite{FD_YOON2018} and facial landmark point detector \cite{DAN_CVPRW2017}. We used $68$ landmark points for the face alignment and extraction of local appearance features. When the detection of faces or facial landmark points is failed, we simply discard the image. Thus, we discarded $24,160$ face images from $6,561$ subjects. After removing these images without landmark points, it roughly goes to $3.1$M images of $8,630$ unique persons. We generated a validation set by selecting randomly about $10\%$ from each subject in refined dataset, and the remains are used as the training set. Therefore, the training set roughly has $2.8$M face images and the validation set has $311,773$ face images, respectively.

\subsection{Data preprocessing}
\begin{figure}[h]
	\centering
	\includegraphics[scale=0.45]{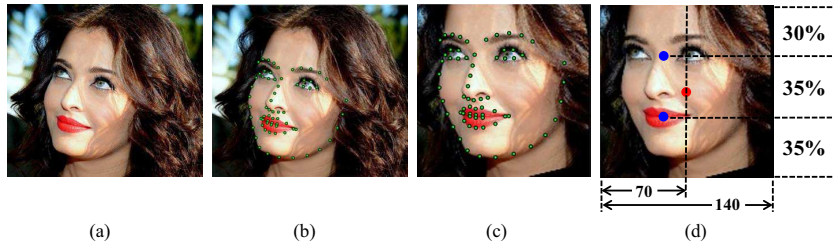}
	\caption{A face image alignment example. The original image is shown in (a); (b) shows the $68$ landmark points detected by the method in \cite{DAN_CVPRW2017}, (c) shows the $68$ landmark points aligned into the aligned image plane; and (d) is the final aligned face image, where the red circle was used to center the face image along $x$-axis, and the blue circles denote the two points used for face cropping.}
	\label{fig:fig_alignment}
\end{figure}

We employ a new face alignment to align training face images into the predefined template. The alignment procedures are as follows: 1) Use the DAN implementation of Kowalski \textit{et al.} by using multi-stage neural network \cite{DAN_CVPRW2017} to detect $68$ facial landmarks (\figurename~\ref{fig:fig_alignment}b); 2) rotate the face in the image to make it upright based on the eye positions; 3) find a center on the face by taking the mid-point between the leftmost and rightmost landmark points (the red point in \figurename~\ref{fig:fig_alignment}d); 4) the centers of the eyes and mouth (blue points in \figurename~\ref{fig:fig_alignment}d) are found by averaging all the landmark points in the eye and mouth regions; 5) center the faces in the $x$-axis, based on the center (red point); 6) fix the position along the $y$-axis by placing the eye center at $30\%$ from the top of the image and the mouth center at $35\%$ from the bottom of the image; 7) resize the image to a resolution of $140\times140$. Each pixel which value is in a range of $[0, 255]$ in the RGB color space is normalized by dividing $255$ to be in a range of $[0, 1]$.

\subsection{Base CNN model}\label{sec:base_cnn}
The base CNN model is the backbone neural network which accepts the RGB values of the aligned face image with $140\times140$ resolution as its input, and has $64$ $5\times5$ convolution filters with a stride of $1$ in the first layer. After $3\times3$ max pooling with a stride of $2$, it has several 3-layer residual bottleneck blocks similar to the ResNet-101 \cite{ResNet}. In the last layer, we use the global average pooling with $9\times9$ filter in each channel and use the fully connected layer. The output of the fully connected layer are fed into \textit{softmax} loss layer (\tablename~\ref{tab:tab_baseline_cnn}).
\begin{table}[h]
	\centering 
	\caption{A backbone conovlutional neural network architecture.}
	\label{tab:tab_baseline_cnn}
	\begin{tabular}{@{}c|c|c@{}}
		\toprule
		\textbf{Layer name}    & \textbf{Output size} & \textbf{101-layer} \\ \hline
		\textit{conv1}         & $140\times140$       & $5\times5$, $64$         \\ \hline
		\multirow{2}{*}{\textit{conv2\_x}} & \multirow{2}{*}{$70\times70$} & $3\times3$ max pool, stride $2$       \\ \cline{3-3}
		&                      & $\left[\begin{array}{c} 1\times1, 64 \\ 3\times3, 64 \\ 1\times1, 256 \end{array} \right] \times3$     \\ \hline
		\textit{conv3\_x}      & $35\times35$         & $\left[\begin{array}{c} 1\times1, 128 \\ 3\times3, 128 \\ 1\times1, 512 \end{array} \right] \times4$   \\ \hline
		\textit{conv4\_x}      & $18\times18$         & $\left[\begin{array}{c} 1\times1, 256 \\ 3\times3, 256 \\ 1\times1, 1024 \end{array} \right] \times23$  \\ \hline
		\textit{conv5\_x}      & $9\times9$           & $\left[\begin{array}{c} 1\times1, 512 \\ 3\times3, 512 \\ 1\times1, 2048 \end{array} \right] \times3$  \\ \hline
		& $1\times1$           &  average pool, 8630-d fc, \textit{softmax}         \\ \bottomrule
	\end{tabular}
\end{table}

\subsection{Detailed settings in the PRN}\label{sec:implement_details_prn}
For pairwise relations between facial parts, we first extracted a set of local appearance feature $\boldsymbol{F}^{l}$ from each local region (nearly $1\times1$ size of regions) around $68$ landmark points by ROI projection on the $9\times9\times2,048$ feature maps (\textit{conv5\_3} in \tablename~\ref{tab:tab_baseline_cnn}) in the backbone CNN model. Using this $\boldsymbol{F}^{l}$, we make $2,278$ ($={}^{68}C_{2}$) possible pairs of local appearance features. Then, we used three-layered MLP consisting of $1,000$ units per layer with BN and ReLU non-linear activation functions for $\mathcal{G}_{\theta}$, and three-layered MLP consisting of $1,000$ units per layer with BN and ReLU non-linear activation functions for $\mathcal{F}_{\phi}$. To aggregate all of relations from $\mathcal{G}_{\theta}$, we used summation as an aggregation function.
The PRN is optimized jointly with \textit{triplet ratio} loss, \textit{pairwise} loss, and \textit{softmax} loss over the ground-truth identity labels using stochastic gradient descent (SGD) optimization method with learning rate $0.10$.
We used $128$ mini-batches size on four NVIDIA Titan X GPUs. During training the PRN, we froze the backbone CNN model to only update weights of the PRN model.

\subsection{Face identity state feature}\label{sec:sid_config}
\begin{figure}[t]
	\centering
	\includegraphics[scale=0.43]{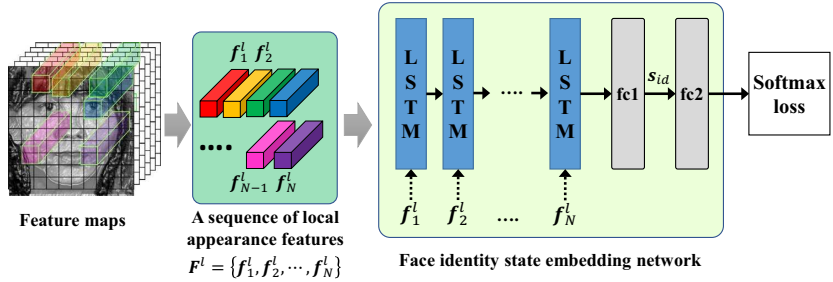}
	\caption{Face identity state feature. A face on the feature maps is divided into $68$ regions by ROI projection around $68$ landmark points. A sequence of local appearance features in these regions are used to encode the face identity state feature from LSTM networks.}
	\label{fig:fig_idf}
\end{figure}

Pairwise relations should be identity dependent to capture unique pairwise relations within same identity and discriminative pairwise relations between different identities.
Based on the feature maps in the CNN, the face is divided into $68$ local regions by ROI projection around $68$ landmark points. In these local regions, we extract the local appearance features to encode the facial identity state feature $\boldsymbol{s}_{id}$.
Let $\boldsymbol{f}^{l}_{i}$ denote the local appearance feature of $m^{'}\times m^{'}$  $i$-th local region. To encode $\boldsymbol{s}_{id}$, an LSTM-based network has been devised on top of a set of local appearance features $\boldsymbol{F}^{l} = \{\boldsymbol{f}^{l}_{1}, \ldots, \boldsymbol{f}^{l}_{i}, \ldots, \boldsymbol{f}^{l}_{N}\}$ as followings:
\begin{equation}
\boldsymbol{s}_{id} = \mathcal{E}_{\psi}(\boldsymbol{F}^{l}), 
\label{eq:equation_lstm}
\end{equation}
where $\mathcal{E}_{\psi}$ is a neural network module which composed of LSTM layers and two fully connected layers with learnable parameters $\psi$. We train $\mathcal{E}_{\psi}$ with \textit{softmax} loss function (\figurename~\ref{fig:fig_idf}). 
To capture unique and discriminative pairwise relations dependent on identity, the PRN should condition its processing on the face identity state feature $\boldsymbol{s}_{id}$. For $\boldsymbol{s}_{id}$, we use the LSTM-based recurrent network $\mathcal{E}_{\psi}$ over a sequence of the local appearance features which is a set ordered by landmark points order from $\boldsymbol{F}^{l}$. In other words, there were a sequence of $68$ length per face. In $\mathcal{E}_{\psi}$, it consist of LSTM layers and two-layer MLP. Each of the LSTM layer has $2,048$ memory cells. The MLP consists of $256$ and $8,630$ units per layer, respectively. The cross-entropy loss with \textit{softmax} was used for training the $\mathcal{E}_{\psi}$ (\figurename~\ref{fig:fig_idf}).

\subsection{Detailed settings in the model}\label{sec:detail_setting_models}
We implemented the base CNN and the PRN model using the Keras framework \cite{Keras_LIB} with TensorFlow \cite{Tensorflow} backend. For fair comparison in terms of the effects of each network module, we train three kinds of models (\textbf{model A}, \textbf{model B}, and \textbf{model C}) under the supervision of cross-entropy loss with \textit{softmax} \cite{PIMNet_CVPRW2017}:
\textbf{model A} is the baseline model which is the base CNN (Table \ref{tab:tab_baseline_cnn}).
\textbf{model B} combining two different networks, one of which is the base CNN model (\textbf{model A}) and the other is the $PRN$ (Eq. \eqref{eq:eq_prn_rev}), concatenates the output feature $\boldsymbol{f}^{g}$ of the global average pooling layer in \textbf{model A} as the global appearance feature and the output of the MLP $\mathcal{F}_{\phi}$ in the $PRN$. $\boldsymbol{f}^{g}$ is the feature of size $1\times1\times2,048$ from each face image. The output of the MLP $\mathcal{F}_{\phi}$ in the $PRN$ is the feature of size $1\times1\times1,000$. These two output features are concatenated into a single feature vector with $3,048$ size, then this feature vector is fed into the fully connected layer with $1,024$ units. \textbf{model C} is the combined model with the output of \textbf{model A} and the the output of the $PRN^{+}$ (Eq. \eqref{eq:equation_prn_idf}). The output of \textbf{model A} in \textbf{model C} is the same of the output in \textbf{model B}. The size of the output in the $PRN^{+}$ is same as compared with the ${PRN}$, but output values are different.
All of convolution layers and fully connected layers use batch normalization (BN) \cite{BatchNormal} and rectified linear units (ReLU) \cite{ReLU} as nonlinear activation functions except for LSTM laeyrs in $\mathcal{E}_{\psi}$.

\section{Detailed Results}
\subsection{Experiments on the LFW}\label{sec:lfw_appendix}
We evaluated the proposed method on the LFW dataset \cite{LFW}, which reveals the \textit{state-of-the-art} of face verification in unconstrained environments. LFW dataset is excellent benchmark dataset for face verification in image and contains $13,233$ web crawling images with large variations in illuminations, occlusions, facial poses, and facial expressions, from $5,749$ different identities.
Our models such as \textbf{model A}, \textbf{model B}, and \textbf{model C} were trained on the roughly $2.8$M outside training set, with no people overlapping with subjects in the LFW. Following the test protocol of \textit{unrestricted with labeled outside data} \cite{LFWTechUpdate}, we test on $6,000$ face pairs by using a squared $L_{2}$ distance threshold to determine classification of \textit{same} and \textit{different} and report the results in comparison with the \textit{state-of-the-art} methods (\tablename~\ref{tab:lfw_results}).

\begin{table}[h]
	\small
	\centering
	\caption{Comparison of the number of images, the number of networks, the dimensionality of feature, and the accuracy of the proposed method with the \textit{state-of-the-art} methods on the LFW.}
	\label{tab:lfw_results}
	\resizebox{12.3cm}{!} {
		\begin{tabular}{lcccc}
			\toprule
			\multicolumn{1}{c}{\textbf{Method}}   							& \textbf{~~Images~~} & \textbf{~~Networks~~} & \textbf{~~Dimension~~} & \textbf{~~Accuracy (\%)~~} \\
			\hline
			Human                                 							& -                    & -                      & -                          & $97.53$          \\
			DeepFace \cite{DeepFace}              							& $4$M				   & $9$                    & $4,096\times4$             & $97.25$          \\
			DeepID \cite{DeepID}                  							& $202,599$			   & $120$                  & $150\times120$             & $97.45$          \\
			DeepID2+ \cite{DeepID2}               							& $300,000$			   & $25$                   & $150\times120$             & $99.47$          \\
			DeepID3 \cite{DeepID3}                							& $300,000$ 		   & $50$                   & $300\times100$             & $99.52$          \\
			FaceNet \cite{FaceNet}                							& $200$M		   	   & $1$                    & $128$                      & $99.63$          \\
			Learning from Scratch \cite{LFR}      							& $494,414$			   & $2$                    & $160\times2$               & $97.73$          \\
			Center Face \cite{CenterLoss}         							& $0.7$M			   & $1$                    & $512$                      & $99.28$          \\
			PIMNet${}_{\textrm{TL-Joint~Bayesian}}$ \cite{PIMNet_CVPRW2017} & $198,018$			   & $4$                    & $1,024$                    & $98.33$ \\
			PIMNet${}_{\textrm{fusion}}$ \cite{PIMNet_CVPRW2017}            & $198,018$			   & $4$                    & $6$                        & $99.08$  \\
			SphereFace \cite{SphereFace}          							& $494,414$			   & $1$                    & $1024$                     & $99.42$          \\
			ArcFace \cite{ArcFace}                							& $3.1$M			   & $1$                    & $512$                      & $99.78$          \\ \midrule
			\textbf{model A} (baseline)            							& $2.8$M			   & $1$                    & $2,048$                    & $\mathbf{99.6}$     \\
			\textbf{model B}                      							& $2.8$M			   & $1$                    & $1,000$                    & $\mathbf{99.65}$    \\
			\textbf{model C}                      							& $2.8$M			   & $1$                    & $1,024$                    & $\mathbf{99.76}$    \\ \bottomrule
		\end{tabular}
	}
\end{table}

\subsection{Experiments on the YTF}\label{sec:ytf_appendix}
We evaluated the proposed method on the YTF dataset\cite{YTF_CVPR2011}, which reveals the \textit{state-of-the-art} of face verification in unconstrained environments. YTF dataset is excellent benchmark dataset for face verification in video and contains $3,425$ videos with large variations in illuminations, facial pose, and facial expressions, from $1,595$ different identities, with an average of $2.15$ videos per person. The length of video clip varies from $48$ to $6,070$ frames and average of $181.3$ frames. We follow the test protocol of \textit{unrestricted with labeled outside data}. We test on $5,000$ video pairs by using a squared $L_{2}$ distance threshold to determine to classification of \textit{same} and \textit{different} and report the results in comparison with the \textit{state-of-the-art} methods (Table \ref{tab:YTF_results}).

\begin{table}[h]
	\small
	\centering
	\caption{Comparison of the number of CNNs, the number of images, the dimensionality of feature, and the accuracy of the proposed method with the \textit{state-of-the-art} methods on the YTF.}
	\label{tab:YTF_results}
	\resizebox{12.5cm}{!} {
		\begin{tabular}{lcccc}
			\toprule
			\multicolumn{1}{c}{\textbf{Method}}   & \textbf{~~Images~~}  & \textbf{~~Networks~~}  & \textbf{~~Dimension~~}  & \textbf{~~Accuracy (\%)~~} \\
			\hline
			DeepFace \cite{DeepFace}              & $4$M                 & $9$                    & $4,096\times4$          & $91.4$          \\
			DeepID2+ \cite{DeepID2}               & $300,000$            & $25$                   & $150\times120$          & $93.2$          \\
			FaceNet \cite{FaceNet}                & $200$M               & $1$                    & $128$                   & $95.1$          \\
			Learning from Scratch \cite{LFR}      & $494,414$            & $2$                    & $160\times2$            & $92.2$          \\
			Center Face \cite{CenterLoss}         & $0.7$M               & $1$                    & $512$                   & $94.9$          \\
			SphereFace \cite{SphereFace}          & $494,414$            & $1$                    & $1024$                  & $95.0$          \\
			NAN \cite{NAN_CVPR2017}               & $3$M                 & $1$                    & $128$                   & $95.7$          \\ \midrule
			\textbf{model A} (baseline)           & $2.8$M               & $1$                    & $2,048$                 & $\mathbf{95.1}$     \\
			\textbf{model B}                      & $2.8$M               & $1$                    & $1,000$                 & $\mathbf{95.7}$    \\
			\textbf{model C}                      & $2.8$M               & $1$                    & $1,024$                 & $\mathbf{96.3}$    \\ \bottomrule
		\end{tabular}
	}
\end{table}

\subsection{Experiments on the IJB-A}\label{sec:ijb-a_appendix}
We evaluated the proposed method on the IJB-A dataset \cite{IJB-A_CVPR2015} which contains face images and videos captured from unconstrained environments. It features full pose variation and wide variations in imaging conditions thus is very challenging. It contains $500$ subjects with $5,397$ images and $2,042$ videos in total, and $11.4$ images and $4.2$ videos per subject on average. In this dataset, each training and testing instance is called a 'template', which comprises $1$ to $190$ mixed still images and video frames. IJB-A dataset provides $10$ split evaluations with two protocols (1:1 face verification and 1:N face identification). For face verification, we report the test results by using true accept rate (TAR) \textit{vs.} false accept rate (FAR) (\tablename~\ref{tab:ijb-a_results}). For face identification, we report the results by using the true positive identification (TPIR) \textit{vs.} false positive identification rate (FPIR) and Rank-N (\tablename~\ref{tab:ijb-a_results}). All measurements are based on a squared $L_{2}$ distance threshold.

\begin{table}[h]
	\centering
	\caption{Comparison of performances of the proposed PRN method with the state-of-the-art on the IJB-A dataset. For verification, the true accept rates (TAR) \textsl{vs.} false accept rates (FAR) are reported. For identification, the true positive identification rate (TPIR) \textsl{vs.} false positive identification rate (FPIR) and the Rank-N accuracies are presented.}
	\label{tab:ijb-a_results}
	\resizebox{\textwidth}{!} {
		\begin{tabular}{@{}llllllllll@{}}
			\toprule
			\multicolumn{1}{c}{\multirow{2}{*}{Method}}           & \multicolumn{3}{c}{1:1 Verification TAR} & \multirow{2}{*}{} & \multicolumn{5}{c}{1:N Identification TPIR}  \\ \cmidrule(lr){2-4} \cmidrule(l){6-10}
			\multicolumn{1}{c}{}                                  & FAR=0.001        &  FAR=0.01         &  FAR=0.1          &  &  FPIR=0.01        &  FPIR=0.1         &  Rank-1           & Rank-5           &  Rank-10           \\ \midrule
			B-CNN \cite{B-CNN_WACV2016}                           & -                &  -                &  -                &  &  $0.143\pm0.027$  &  $0.341\pm0.032$  &  $0.588\pm0.020$  &  $0.796\pm0.017$  &  -                \\
			LSFS \cite{LSFS_PAMI2017}                             & $0.514\pm0.060$  &  $0.733\pm0.034$  &  $0.895\pm0.013$  &  &  $0.383\pm0.063$  &  $0.613\pm0.032$  &  $0.820\pm0.024$  &  $0.929\pm0.013$  &  -                \\
			DCNN${}_{manual}$+metric \cite{DCNN_Metric_ICCVW2015} & -                &  $0.787\pm0.043$  &  $0.947\pm0.011$  &  &  -                &  -                &  $0.852\pm0.018$  &  $0.937\pm0.010$  &  $0.954\pm0.007$  \\
			Triplet Similarity \cite{TripletSimilarity_BTAS2016}  & $0.590\pm0.050$  &  $0.790\pm0.030$  &  $0.945\pm0.002$  &  &  $0.556\pm0.065$  &  $0.754\pm0.014$  &  $0.880\pm0.015$  &  $0.95\pm0.007$   &  $0.974\pm0.005$  \\
			Pose-Aware Models \cite{POSE_AWARE_CVPR2016}          & $0.652\pm0.037$  &  $0.826\pm0.018$  &  -                &  &  -                &  -                &  $0.840\pm0.012$  &  $0.925\pm0.008$  &  $0.946\pm0.005$  \\
			Deep Multi-Pose \cite{Deep_Multi_Pose_WACV2016}       & -                &  $0.876$          &  $0.954$          &  &  $0.52$           &  $0.75$           &  $0.846$          &  $0.927$          &  $0.947$          \\
			DCNN${}_{fusion}$ \cite{DCNN_Fusion_WACV2016}         & -                &  $0.838\pm0.042$  &  $0.967\pm0.009$  &  &  $0.577\pm0.094$  &  $0.790\pm0.033$  &  $0.903\pm0.012$  &  $0.965\pm0.008$  &  $0.977\pm0.007$  \\
			Triplet Embedding \cite{TripletSimilarity_BTAS2016}   & $0.813\pm0.02$   &  $0.90\pm0.01$    &  $0.964\pm0.005$  &  &  $0.753\pm0.03$   &  $0.863\pm0.014$  &  $0.932\pm0.01$   &  -                &  $0.977\pm0.005$  \\
			VGG-Face \cite{VGGFACE_BMVC2015}                      & -                &  $0.805\pm0.030$  &  -                &  &  $0.461\pm0.077$  &  $0.670\pm0.031$  &  $0.913\pm0.011$  &  -                &  $0.981\pm0.005$  \\
			Template Adaptation \cite{Template_ADAPT_FGR2017}     & $0.836\pm0.027$  &  $0.939\pm0.013$  &  $0.979\pm0.004$  &  &  $0.774\pm0.049$  &  $0.882\pm0.016$  &  $0.928\pm0.010$  &  $0.977\pm0.004$  &  $0.986\pm0.003$  \\
			NAN \cite{NAN_CVPR2017}                               & $0.881\pm0.011$  &  $0.941\pm0.008$  &  $0.978\pm0.003$  &  &  $0.817\pm0.041$  &  $0.917\pm0.009$  &  $0.958\pm0.005$  &  $0.980\pm0.005$  &  $0.986\pm0.003$  \\
			VGGFace2 \cite{VGG2Face}                              & $0.921\pm0.014$  &  $0.968\pm0.006$  &  $0.990\pm0.002$  &  &  $0.883\pm0.038$  &  $0.946\pm0.004$  &  $0.982\pm0.004$  &  $0.993\pm0.002$  &  $0.994\pm0.001$  \\ \midrule
			\textbf{model A} (baseline)                           & $\mathbf{0.895\pm0.015}$  &  $\mathbf{0.949\pm0.008}$  &  $\mathbf{0.980\pm0.005}$  &  &  $\mathbf{0.843\pm0.035}$  &  $\mathbf{0.923\pm0.005}$  &  $\mathbf{0.975\pm0.005}$  &  $\mathbf{0.992\pm0.004}$  &  $\mathbf{0.993\pm0.001}$  \\
			\textbf{model B}                                      & $\mathbf{0.901\pm0.014}$  &  $\mathbf{0.950\pm0.006}$  &  $\mathbf{0.985\pm0.002}$  &  &  $\mathbf{0.861\pm0.038}$  &  $\mathbf{0.931\pm0.004}$  &  $\mathbf{0.976\pm0.003}$  &  $\mathbf{0.992\pm0.003}$  &  $\mathbf{0.994\pm0.003}$  \\
			\textbf{model C}                                      & $\mathbf{0.919\pm0.013}$  &  $\mathbf{0.965\pm0.004}$  &  $\mathbf{0.988\pm0.002}$  &  &  $\mathbf{0.882\pm0.038}$  &  $\mathbf{0.941\pm0.004}$  &  $\mathbf{0.982\pm0.004}$  &  $\mathbf{0.992\pm0.002}$  &  $\mathbf{0.995\pm0.001}$  \\ \bottomrule
		\end{tabular}
	}
\end{table}

\subsection{Experiments on the IJB-B}\label{sec:ijb-b_appendix}
We evaluated the proposed method on the IJB-B dataset \cite{IJB-B_CVPRW2017} which contains face images and videos captured from unconstrained environments.
The IJB-B dataset is an extension of the IJB-A, having $1,845$ subjects with $21.8$K still images (including $11,754$ face and $10,044$ non-face) and $55$K frames from $7,011$ videos, an average of $41$ images per subject.
Because images in this dataset are labeled with ground truth bounding boxes, we only detect landmark points using DAN \cite{DAN_CVPRW2017}, and then align face images with our face alignment method.
Unlike the IJB-A, it does not contain any training splits. In particular, we use the 1:1 Baseline Verification protocol and 1:N Mixed Media Identification protocol for the IJB-B. For face verification, we report the test results by using TAR \textit{vs.} FAR (\tablename~\ref{tab:ijb-b_results}). For face identification, we report the results by using TPIR \textit{vs.} FPIR and Rank-N (\tablename~\ref{tab:ijb-b_results}).
We compare our proposed methods with VGGFace2 \cite{VGG2Face} and FacePoseNet (FPN) \cite{FPN_Align_ICCVW2017}. All measurements are based on a squared $L_{2}$ distance threshold.

\begin{table}[h]
	\small
	\centering
	\caption{Comparison of performances of the proposed PRN method with the \textit{state-of-the-art} on the IJB-B dataset. For verification, TAR \textsl{vs.} FAR are reported. For identification, TPIR \textsl{vs.} FPIR and the Rank-N accuracies are presented}
	\label{tab:ijb-b_results}
	\resizebox{\textwidth}{!} {
		\begin{tabular}{@{}lllllllllll@{}}
			\toprule
			\multicolumn{1}{c}{\multirow{2}{*}{Method}} & \multicolumn{4}{c}{1:1 Verification TAR} & \multirow{2}{*}{} & \multicolumn{5}{c}{1:N Identification TPIR}  \\ \cmidrule(lr){2-5} \cmidrule(l){7-11}
			\multicolumn{1}{c}{}                        			& FAR=0.00001  &  FAR=0.0001  &  FAR=0.001  &  FAR=0.01  &  &  FPIR=0.01  &  FPIR=0.1  &  Rank-1  &  Rank-5  &  Rank-10  \\ \midrule
			VGGFace2 \cite{VGG2Face}                    			& $0.671$  & $0.800$  & $0.0.888$  & $0.949$  &  &  $0.746\pm0.018$  &  $0.842\pm0.022$  &  $0.912\pm0.017$  &  $0.949\pm0.010$  &  $0.962\pm0.007$  \\
			VGGFace2\_ft \cite{VGG2Face}                			& $0.705$  & $0.831$  & $0.908$    & $0.956$  &  &  $0.763\pm0.018$  &  $0.865\pm0.018$  &  $0.914\pm0.029$  &  $0.951\pm0.013$  &  $0.961\pm0.010$  \\
			FPN \cite{FPN_Align_ICCVW2017}              			& -  	   & $0.832$  & $0.916$    & $0.965$  &  &  -  				 &  -  				 &  $0.911$  		 &  $0.953$  		 &  $0.975$  		\\ \midrule
			\textbf{model A} (baseline, only $\boldsymbol{f}^{g}$)  & $\mathbf{0.673}$  & $\mathbf{0.812}$  &  $\mathbf{0.892}$  &  $\mathbf{0.953}$  &  &  $\mathbf{0.743\pm0.019}$  &  $\mathbf{0.851\pm0.017}$  &  $\mathbf{0.911\pm0.017}$  &  $\mathbf{0.950\pm0.013}$  &  $\mathbf{0.961\pm0.010}$  \\
			\textbf{model B} ($\boldsymbol{f}^{g}$ + $PRN$)         & $\mathbf{0.692}$  & $\mathbf{0.829}$  &  $\mathbf{0.910}$  &  $\mathbf{0.956}$  &  &  $\mathbf{0.773\pm0.018}$  &  $\mathbf{0.865\pm0.018}$  &  $\mathbf{0.913\pm0.022}$  &  $\mathbf{0.954\pm0.010}$  &  $\mathbf{0.965\pm0.013}$  \\
			\textbf{model C} ($\boldsymbol{f}^{g}$ + $PRN^{+}$)     & $\mathbf{0.721}$  & $\mathbf{0.845}$  &  $\mathbf{0.923}$  &  $\mathbf{0.965}$  &  &  $\mathbf{0.814\pm0.017}$  &  $\mathbf{0.907\pm0.013}$  &  $\mathbf{0.935\pm0.015}$  &  $\mathbf{0.965\pm0.017}$  &  $\mathbf{0.975\pm0.007}$  \\ \bottomrule
		\end{tabular}
	}
\end{table}

\end{document}